\documentclass{article}
\usepackage{spconf,amsmath,graphicx}

\usepackage{xcolor,colortbl}
\usepackage{hyperref}
\usepackage{graphicx}
\usepackage{adjustbox}
\usepackage{amssymb}
\usepackage{pifont}
\usepackage{multirow}
\usepackage{amsmath}
\usepackage{bbm}
\usepackage{float}
\usepackage{caption}
\usepackage{subcaption}
\usepackage{booktabs}

\usepackage[para,online,flushleft]{threeparttable}

\definecolor{cinnabar}{rgb}{0.89, 0.26, 0.2}
\definecolor{citrine}{rgb}{0.89, 0.82, 0.04}
\definecolor{cobalt}{rgb}{0.0, 0.28, 0.67}
\definecolor{darkorchid}{rgb}{0.6, 0.2, 0.8}
\definecolor{darkpastelgreen}{rgb}{0.01, 0.75, 0.24}

\usepackage[english]{babel}

\title{A Focused Study on Sequence Length for Dialogue Summarization}

\name{
Bin Wang \textsuperscript{\rm \dag},
Chen Zhang \textsuperscript{\rm \dag},
Chengwei Wei \textsuperscript{\rm \S},
Haizhou Li \textsuperscript{\rm \ddag, \rm \dag, \rm $\sharp$}
}
\address{
\textsuperscript{\rm \dag} National University of Singapore, Singapore
\quad
\textsuperscript{\rm \S} University of Southern California, USA
\\
\textsuperscript{\rm \ddag} The Chinese University of Hong Kong, Shenzhen, China \quad
\quad
\textsuperscript{\rm $\sharp$} Kriston AI, China
\\
\texttt{bwang28c@gmail.com}
}

\begin{document}

\maketitle

\begin{abstract}

    Output length is critical to dialogue summarization systems. The dialogue summary length is determined by multiple factors, including dialogue complexity, summary objective, and personal preferences. In this work, we approach dialogue summary length from three perspectives. First, we analyze the length differences between existing models' outputs and the corresponding human references and find that summarization models tend to produce more verbose summaries due to their pretraining objectives. Second, we identify salient features for summary length prediction by comparing different model settings. Third, we experiment with a length-aware summarizer and show notable improvement on existing models if summary length can be well incorporated. Analysis and experiments are conducted on popular DialogSum and SAMSum datasets to validate our findings.\footnote{Code available: \url{https://github.com/BinWang28/LA-BART}}

\end{abstract}


\section{Introduction}

    Previous work in summarization focus on the news domain where both extractive and abstractive methods are effective \cite{el2021automatic}. Dialogue summarization aims to generate summaries for intercourses between individuals or groups. Because of the verbal or written format of dialogues, its summarization is abstractive in nature and poses unique challenges \cite{sacks1978simplest,wang2022analyzing}. A good dialogue summary should be not only coherent and fluent but also concise as well as cover essential details \cite{chen2020multi}. 
    
    Two main factors determine the desired summary length: 1) dialogue complexity and 2) user preferences. Traditional text summarization systems require both source text and summary length as the input to satisfy specific needs like display constrain \cite{kikuchi2016controlling}. In comparison, recent text summarizers, especially pre-trained models, generate a summary without explicit length modeling procedures \cite{chen2020multi,lewis2019bart,liu-etal-2021-coreference}. However, the output sequence length for dialogue summarization is a crucial factor since a concise and informative summary is favorable in both applications and evaluations. The study of performance evaluation, predictability and controllability on the length of dialogue summary is missing.

    \begin{table}[t]
      \centering
      \begin{adjustbox}{width=0.45\textwidth,center}
        \begin{tabular}{ | l | }
            \hline
            \emph{\textbf{Dialogue}}: \\ \hline
            
            \textcolor{darkorchid}{\emph{\#Person1\#}}: May, do you mind helping me prepare for the picnic? \\ 
            \textcolor{darkorchid}{\emph{\#Person2\#}}: Sure. Have you checked the weather report? \textcolor{darkorchid}{\emph{\#Person1\#}}: \\
            Yes. It says it will be sunny all day. No sign of rain at all. This is \\ 
            your father's favorite sausage. Sandwiches for you and Daniel. \\
            \textcolor{darkorchid}{\emph{\#Person2\#}}: No, thanks Mom. I'd like some . . . . . . \\ \hline
            
            \emph{\textbf{Summaries}}: \\\hline
            
            \textcolor{teal}{\emph{Human1}}: May is helping her mother to do some preparation for the \\ picnic.  \\ \hline
            
            \textcolor{teal}{\emph{Human2}}: Mom asks May to help to prepare for the picnic and May \\ agrees. \\ \hline
            
            \textcolor{teal}{\emph{Human3}}: May's mother asks May for help in preparing for a picnic. \\
            May gives her a hand.  \\ \hline
            
            \textcolor{teal}{\emph{BART\textsubscript{Large}}}: \emph{\#Person1\#} asks May to help her prepare for the picnic. \\
            May takes some fruit salad, crackers, sausage, toast, chicken \\ 
            wings, napkins, cups, and picnic blanket to the living room. \\ \hline
            
        \end{tabular}
      \end{adjustbox}
      \caption{One sample from DialogSum dataset.}
      \label{tab:example} 
    \end{table}

    \begin{table*}[t]
        \centering
        \begin{adjustbox}{width=0.87\textwidth,center}
        \begin{tabular}{ l  c c c  c c c  c c c  c}
        \toprule
        \multirow{2}{*}{\textbf{Model}} & \multicolumn{3}{c}{\textbf{ROUGE-1}}              & \multicolumn{3}{c}{\textbf{ROUGE-2}}          & \multicolumn{3}{c}{\textbf{ROUGE-L}} & \multirow{2}{*}{\textbf{Len. $\Delta$}} \\
        
         & \textbf{Prec.} & \textbf{Rec.} & \textbf{F-1} & \textbf{Prec.} & \textbf{Rec.} & \textbf{F-1} & \textbf{Prec.} & \textbf{Rec.} & \textbf{F-1} & \\
        \hline  
        \emph{BART\textsubscript{Large}} & 44.11 & 54.25 & 47.22 & 19.86 & 23.78 & 20.92 & 41.97 & 49.77 & 44.57 & 7.9 \\
        
        \emph{BART\textsubscript{Base}} & 45.87 & 49.10 & 46.12 & 19.78 & 20.71 & 19.63 & 43.43 & 45.97 & 43.81 & 5.4 \\
        
        \emph{T5\textsubscript{Base}} & 40.91 & 48.61 & 43.25 & 17.06 & 19.58 & 17.72 & 38.85 & 44.89 & 40.86 & 6.6 \\
        
        \emph{T5\textsubscript{Small}} & 38.30 & 42.64 & 39.12 & 14.44 & 15.44 & 14.40 & 36.77 & 40.14 & 37.55  & 6.2 \\
        \hline
        \rowcolor{gray!5} Inter-human & 54.16 & 54.68 & 53.34 & 27.17 & 27.35 & 26.70 & 51.38 & 51.78 & 50.84 & 4.2 \\
        \bottomrule
        \end{tabular}
        \end{adjustbox}
        \caption{ROUGE (Precision, Recall, F-1) scores on DialogSum dataset and the absolute length difference between summaries. Inter-human scores are the average between any two annotators.}
        \label{tab:rouge_comparison}
    \end{table*}

    For length-controllable summarization, \cite{kikuchi2016controlling} and \cite{fan2017controllable} propose to use length embedding as input to the decoder of LSTM-based or CNN-based models. \cite{saito2020length} suggests controlling the summary length by first extracting salient tokens from the original text. In comparison, our goal is not to propose a better controllable model but a study on the importance of summary lengths. No prior work focuses on summary length analysis for dialogue summarization systems.

    With the availability of a multi-reference dialogue summarization dataset \cite{chen2021dialsumm}, we initiate the study of dialogue summary length while referring to cross-human statistics. First, the SOTA summarization models are compared to humans, and we found that most models are experiencing a verbose problem due to their pretraining objectives. Human still achieves a much higher length agreement compared to machines. Second, we investigate whether summary length is predictable and what information is critical for length prediction. Last, we adapt existing models to length controllable ones by simply adding output length as an additional input. Through experiments, we witness that significant improvements can be achieved with reference summary length. It means that better consideration of summary lengths has significant potential improvement for existing models.

    \begin{table}[t]
        \centering
        \begin{adjustbox}{width=0.30\textwidth,center}
        \begin{tabular}{ l c c c }
        \toprule
        \multirow{2}{*}{\textbf{Model}}  & \multicolumn{3}{c}{\textbf{Correlation}}   \\
         & $r$ & $\rho$ & $\tau$  \\
        \hline 
        \emph{BART\textsubscript{Large}} & 74.6 & 72.9 & 54.6  \\
        
        \emph{BART\textsubscript{Base}} & 72.7 & 71.8 & 54.4 \\
        
        \emph{T5\textsubscript{Base}} & 71.4 & 72.1 & 54.3 \\
        
        \emph{T5\textsubscript{Small}} & 69.6 & 69.6 & 51.0 \\
        \hline
        \rowcolor{gray!5} Inter-human  & 76.9 & 75.6 & 58.6 \\ 
        \bottomrule
        \end{tabular}
        \end{adjustbox}
        \caption{Correlation of summary lengths.}
        \label{tab:len_corr}
    \end{table}

\section{Performance of Existing Models}
    \label{sec:perform_mo}

    \noindent\textbf{Experimental setup}: DialogSum is a recently released dialogue summarization dataset \cite{chen2021dialsumm}. It consists of 13,460 everyday dialogues, which are divided into three subsets: train (12,460), validation (500) and test (500). Each dialogue is associated with one reference summary except for the test set, where three reference summaries are given. An example is shown in Table~\ref{tab:example} with references and one model output. In this section, we focus on DialogSum because its multi-reference summaries allow us to study inter-human statistics. We compute inter-human performance by comparing summaries from any two annotators, and the average is reported. Four pre-trained generation models are selected for experiments including \emph{BART\textsubscript{Large / Base}} \cite{lewis2019bart} and \emph{T5\textsubscript{Base / Small}} \cite{raffel2019exploring}. We did not include the dedicated dialogue summarization models \cite{chen2020multi,liu-etal-2021-coreference,liu-etal-2021-topic-aware} because close performance is witnessed compared with \emph{BART\textsubscript{Large}}.
    
    \noindent\textbf{Our Findings}: Table~\ref{tab:rouge_comparison} shows model performances on ROUGE scores \cite{lin2004rouge}\footnote{`py-rouge' implementation used in this paper for fair comparison~\cite{feng2021survey}.} and the absolute length difference between summaries concerning each dialogue. Summary length is counted by the number of words split by white space. As expected, an obvious gap still exists between automatic summarizers and humans. There are several major differences. First, humans achieve an almost perfect balance between precision and recall in all ROUGE scores. In contrast, all neural summarizers achieve higher recall than precision, which indicates that the summarization models have better coverage of content presented in the reference summary but tend to generate more unnecessary phrases. More specifically, the recall score of ROUGE-1 on \emph{BART\textsubscript{Large}} is extremely close to inter-human performance (54.25 to 54.68). Yet, a lower precision leads to a lower F-1 score. Second, we witness that the absolute length differences for summarizers are larger than for humans. The agreement between humans is a 4.2-word length difference. In comparison, existing summarizers are less competent in determining the summary length and tend to generate longer summaries than desired. Third, Table~\ref{tab:len_corr} shows the length correlation across humans and models in all dialogues. We found that summarizers have comparable performance with humans. It indicates that the summarizer can determine the relative summary length well. In other words, both humans and summarizers can decide which dialogue requires a lengthier summary. 
    
    The above analysis concludes that the existing summarizers suffer from verbose problems. The main reason is from the denoising objective in pretraining. The denoising-based pretraining objective generates the output sequence similar to the input in length \cite{lewis2019bart}, while the summarization task expects a condensed output. Therefore, future pretrained summarization models should pay more attention to verbose problems during unsupervised pretraining or fine-tuning.

    \begin{table}[t]
        \centering
        \begin{adjustbox}{width=0.36\textwidth,center}
        \begin{tabular}{ l c c }
        \toprule
        \textbf{Model}  &  \textbf{DialogSum} & \textbf{SAMSum} \\
        \hline 
        \emph{Surface} & 5.12 & 6.84 \\ \hline
        \emph{Single} & 5.96 & 6.34 \\
        \emph{Single+} & 4.69 & \textbf{6.33} \\ \hline
        \emph{Multi} & 4.83 & 7.19 \\
        \emph{Multi+} & \textbf{4.53} & 7.35 \\
        \hline
        \rowcolor{gray!5} Inter-human & 4.21 & -- \\        
        \bottomrule
        \end{tabular}
        \end{adjustbox}
        \caption{Length prediction results of five models and human performance in absolute length difference.}
        \label{tab:len_pred}
    \end{table}
    
    \begin{table*}[t]
        \centering
        \begin{adjustbox}{width=0.93\textwidth,center}
        \begin{tabular}{ l c c c c c c c c c }
        \toprule
         \multirow{2}{*}{\textbf{Model}} & \multirow{2}{*}{\textbf{Sum. Len.}} & \multicolumn{4}{c}{\textbf{DialogSum}} & \multicolumn{4}{c}{\textbf{SAMSum}} \\
           & & \textbf{R-1} & \textbf{R-2} & \textbf{R-L} & \textbf{Len. $\Delta$} & \textbf{R-1} & \textbf{R-2} & \textbf{R-L} & \textbf{Len. $\Delta$} \\
        \hline
                \multicolumn{10}{l}{\emph{Previous Results}} \\ 
        \hline 
         \emph{Ext-Oracle}  & \ding{55} & 38.68     & 17.28     & 40.06 & - & 42.12     & 17.08     & 40.15  & -   \\
         
         \emph{T5\textsubscript{Base}} & \ding{55} & 43.25 & 17.72 & 40.86 & 6.6 & 52.11 & 26.88 & 49.32 & 10.2 \\
         
         $^\vartriangle$\emph{UniLMv2\textsubscript{Base}}  & \ding{55}  & 47.04     & 21.13     & 45.04 & - & 50.53     & 26.62     & 48.81 & -   \\ 
         
         $^\vartriangle$\emph{BART\textsubscript{Large}} & \ding{55}     & 47.28     & 21.18     & 44.83 & -  & 53.12     & 27.95     & 49.15  & - \\ 
         
         $^\diamond$\emph{MV-BART\textsubscript{Large}}  & \ding{55}    & -        & -         & - & -  & 53.42     & 27.98     & 49.97 & 7.2 \\ 
         
         $^\star$ CODS & \ding{55}   & - & - & - & -   & 52.65 & 27.84 & 50.79 & -  \\
        \hline
        \multicolumn{10}{l}{\emph{Our Results}} \\ 
        \hline
        \emph{BART\textsubscript{Large}} & \ding{55} & 47.22 & 20.92 & 44.57 & 7.9 & 52.93 & 28.21 & 49.88 & 9.3  \\
         \quad +w/ len. out. & \ding{55} & \textbf{47.29} & 20.77 & 45.01 & 5.9  & 53.44 & 28.08 & 50.14 & 8.2 \\
         
        \emph{LA-BART\textsubscript{Large}} & \emph{pseudo} & 47.28 & \textbf{21.09} & \textbf{45.11} & 5.5 & 53.43 & 28.28 & 49.94 & 8.7 \\
         \quad +w/ len. out. & \emph{pseudo} & 47.11 & 20.60 & 44.91 & \textbf{4.9} & \textbf{53.56} & \textbf{28.59} & \textbf{50.29} & \textbf{8.0} \\ \hline

        \emph{LA-BART\textsubscript{Large}} & \ding{51} & \textbf{49.81} & \textbf{22.81} & \textbf{47.40} & 2.6 & 57.81 & 31.73 & 53.46 & 4.1 \\
         \quad +w/ len. out. & \ding{51} & 49.29 & 22.19 & 46.97 & \textbf{1.7} & \textbf{57.89} & \textbf{31.85} & \textbf{53.95} & \textbf{2.7} \\
         
        \bottomrule
        \end{tabular}
        \end{adjustbox}
        \caption{Experimental results on DialogSum and SAMSum datasets. $^\vartriangle$, $^\diamond$, $^\star$ indicate the results are taken from \cite{chen2021dialsumm}, \cite{feng2021survey} and \cite{wu2021controllable}, respectively.}
        \label{tab:main_results}
    \end{table*}

    \begin{table}[t]
        \centering
        \begin{adjustbox}{width=0.4\textwidth,center}
        \begin{tabular}{ l c c c c  }
        \toprule
         \multirow{1}{*}{\textbf{Model}} & \multicolumn{1}{c}{\textbf{DialogSum}} & \multicolumn{1}{c}{\textbf{SAMSum}} \\
        \midrule
        
        \emph{BART\textsubscript{Large}} & 51.6 & \textbf{56.8} \\
         \quad +w/ len. out. & 52.3 & 56.7  \\
         
        \emph{LA-BART\textsubscript{Large}(ps)} & 52.5 & 56.3  \\
         \quad +w/ len. out. & \textbf{52.6} & 56.7  \\ \midrule

        \emph{LA-BART\textsubscript{Large}} & \textbf{54.5} & 59.5   \\
         \quad +w/ len. out. & 54.3 & \textbf{59.8} \\
         
        \bottomrule
        \end{tabular}
        \end{adjustbox}
        \caption{Performance evaluation on BERTScore.}
        \label{tab:berts}
    \end{table}

\section{Summary Length Predictor}
\label{sec:len_pred}

    Previous dialogue summarization models have no length prediction or length control module. Here, we investigate the difficulty of predicting optimal summary length and what kind of information is beneficial in this task. We fine-tune a \emph{T5\textsubscript{Small}} model and design five variants with different input features and training objectives. Experiments are conducted on DialogSum and SAMSum \cite{gliwa2019samsum} datasets. The details of our training objectives are as follows:
    
    \begin{description}
        \item[Surface:] The surface information of a dialogue is used as input.
        
        ``\textit{Length of dialogue: \#\{x\}. Number of utterance: \#\{y\}.}''
        
        \item[Single:] Dialogue is used as input. 
        
        ``\textit{Dialogue: \{D\}.}''
        
        \item[Single+:] Both the surface information and the dialogue are used as input.
        
        ``\textit{Length of dialogue: \#\{x\}. Number of utterance: \#\{y\}. Dialogue: \{D\}.}''
        
        \item[Multi:] The input is the same as \textbf{Single}. The output contains summary length prediction and summary generation in a multi-task learning manner.
        
        \item[Multi+:] The difference from \textbf{Multi} is its input is the same as \textbf{Single+}.
        
    \end{description}

    \begin{table*}[ht]
        \centering
        \begin{adjustbox}{width=0.93\textwidth,center}
        \begin{tabular}{| c | c | c | c | c | c | c |}
        \hline
                 & \emph{Reference 1}  & \emph{Reference 2} & \emph{Reference 3} & \emph{BART} & \emph{LA-BART (pseudo)} & \emph{LA-BART(true)}   \\
        \hline
         Ranking Score & 3.6 & 3.6 & 3.3 & 2.2 & 2.3 & 2.6 \\ 
        \hline 
        \end{tabular}
        \end{adjustbox}
        \caption{Human evaluation results. The higher the better.}
        \label{tab:human_eval}
    \end{table*}

    Table~\ref{tab:len_pred} shows the length prediction results. On DialogSum dataset, surface information serves as a strong baseline and is even better than using the whole dialogue. A further boost is witnessed by combining surface information and dialogue as input. Meantime, multi-task learning brings extra benefits. As a result, \textbf{Multi+} achieves the best performance and is close to human performance with only a 0.3-word difference in length (4.53 v.s. 4.21). Unlike DialogSum, SAMSum dataset has fewer clear instructions in its annotation process. Therefore, the summary length is less consistent and more challenging to forecast. We found that dialogue plays the most important role in length prediction while multi-task learning does not help and is even worse than surface information. In the following section, we choose the best model for each dataset to acquire pseudo summary lengths as the input to the proposed length controllable model.

\section{Length-Aware Model}

    In this section, we first present a simple method to adapt existing models with length controllability. Then, we experiment with several methods to enhance model's length awareness.
    
    \noindent\textbf{Length-aware models}: Unlike previous methods proposed to control the output length through learning additional length embeddings as the input to LSTM-based or CNN-based seq2seq models \cite{kikuchi2016controlling,fan2017controllable}, we directly use the textual description of the desired output length as additional input along with the dialogue as a sign for desired summary length. More specifically, the input is in the form of 
    $$\emph{Summary length: \#\{z\}. Dialogue: \{D\}}.$$
    Results in Table \ref{tab:main_results} show that the simple length controllable adaption is effective with pre-trained neural summarizers. We then experiment with the following three methods to probe models' length-awareness:
    \begin{description}
        \item[\emph{LA-BART\textsubscript{v1}}:] Ground-truth summary length is used during training. Pseudo summary length is used for inference. It is acquired by the length predictor in Section~\ref{sec:len_pred}.
        
        \item[\emph{LA-BART\textsubscript{v2}}:] Instead of the pseudo length as in \emph{v1}, reference summary length is used during inference. In this case, we assume that the ground-truth summary length is revealed. It tests a model's length controllability and provides an upper bound for length-awareness models.
        
        \item[Multi-task learning:] The summarizer is trained to predict the summary length along with summary generation. We expect a model to be more aware of summary length in this multi-task learning objective.
    \end{description}

    We use \emph{BART\textsubscript{Large}} as the baseline for most experiments and provide more baseline results in the Appendix. Recent dialogue summarization methods are listed for comparison \cite{chen2020multi,chen2021dialsumm,wu2021controllable}. Besides the standard ROUGE score for summarization, we also report the performance on BERTScore \cite{bert-score} for a more comprehensive evaluation.

    \noindent\textbf{Results and analysis}: The results on ROUGE and BERTScore are shown in Table~\ref{tab:main_results} and \ref{tab:berts}, respectively. 
    
    First, we show that \emph{LA-BART} with golden summary length significantly improved ROUGE score and BERTScore on both datasets. The absolute length difference is only 2.6- and 4.1-word, which indicates that using reference summary length can control the generated summary length well. The evaluation metrics favor the generated summary to have a close length with the reference summary for balanced precision and recall. It indicates that the performance of summarizers can be largely improved if summary length can be accurately predicted or well incorporated.
    
    Second, we witness less absolute length difference with pseudo summary length as input. On DialogSum, the length difference is improved from 7.9- to 5.5-word. A noticeable improvement is also shown on BERTScore. In contrast, the improvement in length difference is minor for SAMSum, and no improvement is shown on BERTScore. We believe it is because the summary length of SAMSum is more challenging to predict than the DialogSum dataset (as shown in Table~\ref{tab:len_pred}) because the latter gives more explicit instructions in the summary labeling process.
    
    Third, we always observe obvious improvement in predicted summary length with multi-task learning by adding summary length prediction as an additional objective. It indicates that the model can pay more attention to output length. There is also some improvement shown on ROUGE and BERTScore. Especially on DialogSum dataset, the BERTScore improves from 51.6 to 52.3 by adding the multi-task learning objective. Here, we show that a simple length-awareness trick can improve the summarizer's performance. Therefore, future dialogue summarization models should pay more attention to the generated summary length to be closer to human-level performance.

    \noindent
    \textbf{Human analysis}: To accommodate the drawbacks of automatic summarization metrics, we also conduct the human evaluation on dialogue summary pairs from DialogSum dataset. The detail of our comparative evaluation is as follows. First, annotators are presented with the original dialogue and six candidate summaries. Three summaries are from the dataset references which are human-written. The rest three are machine generated-summaries including the model outputs of \emph{BART\textsubscript{Large}}, \emph{LA-BART\textsubscript{Large}} with pseudo length labels as the input and \emph{BART\textsubscript{Large}} with the ground-truth length labels as input. Then, the annotator is asked to perform comparative ranking of the summaries based on the overall quality. The highest-ranked summary gets a score of 5, while the lowest is scored 0. Here, ten dialogues are randomly sampled and each dialogue is evaluated by 4 annotators.

    Human evaluation results are shown in Table~\ref{tab:human_eval}. As expected, the human-written reference summaries receive the highest score. \emph{LA-BART\textsubscript{Large}} with ground-truth length achieves the best score among generated summarizers. \emph{LA-BART\textsubscript{Large}} with the pseudo length label shows comparative performance with the baseline.

\section{Conclusion}

    In this work, we study the summary length of dialogues. We spot that recent dialogue summarizers suffer from verbose problems due to their pretraining objective. We show by experiments that the model's potential can be stimulated by considering the summary length. We hope this work can arise the attention on summary length and facilitate the development of summarization models.



\bibliographystyle{IEEEbib}
\bibliography{custom}

\vfill\pagebreak
\quad
\vfill\pagebreak

\section{Related Work}
    
    
        Dialogue summarization attracts more and more research attention with the availability of large-scale labeled datasets. Several different domains of data can be formatted as dialogues for summarization including meetings, emails, customer services, and chit-chats \cite{carletta2005ami,gliwa2019samsum,feigenblat-etal-2021-tweetsumm-dialog,zhong-etal-2021-qmsum}. Each released dataset has its own summarization objective as introduced in \cite{tuggener-etal-2021-summarizing}. In this work, we focus on DialogSum \cite{chen2021dialsumm} and SAMSum \cite{gliwa2019samsum} datasets because they are the first large-scale general-purpose chit-chat dialogue summarization datasets.
        
    
        Dialogue summarization models are mainly abstractive neural models because of the interactive nature of dialogues. Information is scattered across utterances which pose challenges to summarizing dialogues. \cite{see-etal-2017-get} proposed a pointer-generator network to copy words from the source content through attention mechanisms. \cite{chen2020multi} and \cite{liu2019topic} leverage the topic segmentation and conversational structures to better model information exchange for the encoding process of summarization. \cite{liu-etal-2021-coreference} proposes to improve co-references by adjusting the attention distribution within summarization models. In terms of length control, \cite{kikuchi2016controlling} and \cite{fan2017controllable} propose to use length embedding as input to the decoder of LSTM-based or CNN-based models. \cite{saito2020length} suggests controlling the summary length by first extracting salient tokens from the original text. In comparison, our goal is not to propose a better controllable model but a study on the importance of summary lengths. No prior work focuses on summary length analysis for dialogue summarization systems. We initiate the first cross-human study with the availability of multi-reference dataset.

    Table~\ref{tab:dataset} shows more details of DialogSum and SAMSum datasets. The datasets are composed of dialogues of daily activities and human-written summaries. Each dialogue is associated with one human-written summary except for the test set of DialogSum, where three references are given. More dataset details can be found in \cite{chen2021dialsumm} and \cite{gliwa2019samsum}.

    \begin{table}[h]
        \centering
        \begin{adjustbox}{width=0.5\textwidth,center}
        \begin{tabular}{ c | c | c | c | c }
        \toprule
                            & \# Train  & \# Val    & \# Test   & \# Comp. Rate     \\
        \midrule
         DialogSum          & 12,460    & 1,500     & 1,500     & 17.04\% \\ 
        \midrule
         SAMSum             & 14,731    & 818       & 819       & 21.65\% \\ 
        \bottomrule 
        \end{tabular}
        \end{adjustbox}
        \caption{Statistics of dialogue summarization datasets.}
        \label{tab:dataset}
    \end{table}

\section{More Results and Case Study}

    We show the result with \emph{BART\textsubscript{Base}} as the baseline in Table~\ref{tab:appendix_results}. In general, it shows the same trend with the results of \emph{BART\textsubscript{Large}} as in Table~\ref{tab:main_results} and similar conclusions can be drawn.
    
    Table~\ref{tab:ctrlen_example} is a case study of \emph{LA-BART} on length controllability. By increasing the input length signal, we observe that longer summaries can be generated accordingly, and more dialogue details are conveyed in the generated summary.

\section{Verbose Problem of Pre-trained Models}

    In Section~\ref{sec:perform_mo}, we spot the verbose problem of existing pre-trained generation models when applied to summarization tasks. \emph{BART} and \emph{T5} are two popular pre-trained encoder-decoder models. However, during pre-training, they share a similar input-output length. It leads the model to generate similar length output with the input. In contrast, summarization is an information compression process and the compression rate is usually less than 30\% (Table~\ref{tab:dataset}). It means that the output length is less than 30\% of the corresponding input length. This phenomenon applies to both chit-chat dialogue summarization and other summarization types. Therefore, we think researchers should be aware of the verbose problem and it is also an opportunity to improve current pre-training dialogue summarization models.

    \begin{table*}[t]
        \centering
        \begin{adjustbox}{width=0.90\textwidth,center}
        \begin{tabular}{ l c c c c c c c c c }
        \toprule
         \multirow{2}{*}{\textbf{Model}} & \multirow{2}{*}{\textbf{Sum. Len.}} & \multicolumn{4}{c}{\textbf{DialogSum}} & \multicolumn{4}{c}{\textbf{SAMSum}} \\ 
           & & \textbf{R-1} & \textbf{R-2} & \textbf{R-L} & \textbf{Len. $\Delta$} & \textbf{R-1} & \textbf{R-2} & \textbf{R-L} & \textbf{Len. $\Delta$} \\
        \midrule 

        \emph{\textbf{BART\textsubscript{Base}}} & \ding{55} & 45.12 & 18.80 & 42.75 & 5.9 & 51.39 & \textbf{26.66} & \textbf{48.73} & 9.1 \\
         \quad +w/ len. out. & \ding{55} & 45.27 & 18.76 & 42.99 & 5.7 & 51.19 & 25.87 & 47.88 & \textbf{7.7}  \\
         
        \emph{LA-BART\textsubscript{Base}} & \emph{pseudo} & \textbf{45.48} & 18.93 & 43.14 & \textbf{4.6} & \textbf{51.60} & 26.58 & 48.26 & 7.8 \\
         \quad +w/ len. out. & \emph{pseudo} & 45.41 & \textbf{19.08} & \textbf{43.32} & 4.9 & 51.36 & 26.30 & 47.93 & 7.9 \\ \midrule
         
        \emph{LA-BART\textsubscript{Base}} & \ding{51} & 47.50 & 20.34 & 44.96 & \textbf{1.3} & \textbf{55.87} & \textbf{29.45} & \textbf{51.56} & \textbf{2.2}  \\
         \quad +w/ len. out. & \ding{51} & \textbf{47.61} & \textbf{20.73} & \textbf{45.39} & 1.7 & 55.82 & 29.06 & 51.49 & 2.3 \\
         
        \bottomrule
        
        \end{tabular}
        \end{adjustbox}
        \caption{Experimental results on DialogSum and SAMSum datasets for \emph{BART\textsubscript{Base}}.}
        \label{tab:appendix_results}
    \end{table*}

        \begin{table*}[t]
        \centering
            \centering
            \begin{adjustbox}{width=1.0\textwidth,center}
            \begin{tabular}{ l | l }
            \toprule
            \textbf{Input Dialogue} & 
                            \begin{tabular}{@{}l@{}}
                                \#Person1\#: Happy Birthday, this is for you, Brian. \\ 
                                \#Person2\#: I'm so happy you remember, please come in and enjoy the party. Everyone's here,\\
                                I'm sure you have a good time.\\
                                \#Person1\#: Brian, may I have a pleasure to have a dance with you? \\
                                \#Person2\#: Ok. \\
                                \#Person1\#: This is really wonderful party. \\
                                \#Person2\#: Yes, you are always popular with everyone. and you look very pretty today. \\
                                \#Person1\#: Thanks, that's very kind of you to say. I hope my necklace goes with my dress, and \\
                                 they both make me look good I feel. \\
                                \#Person2\#: You look great, you are absolutely glowing. \\
                                \#Person1\#: Thanks, this is a fine party. We should have a drink together to celebrate your birthday.
                            \end{tabular} \\ \midrule
            \textbf{Reference} &
                            \begin{tabular}{@{}l@{}}
                                \#Person1\# and Brian are at the birthday party of Brian. Brian thinks \#Person1\# looks great and is popular.
                            \end{tabular} \\ \midrule
              \multicolumn{2}{l}{\emph{\textbf{LA-BART}}} \\ \midrule
              $length=5$  &  \#Person1\# celebrates Brian's birthday with him.\\ \midrule
              $length=10$ &  \#Person1\# dances with Brian at his birthday party. \\ \midrule
              $length=15$ &  \#Person1\# celebrates Brian's birthday and dances with him at the party. Brian thinks it's great. \\ \midrule
              $length=20$ &  \#Person1\# wishes Brian a happy birthday and invites him to dance. Brian agrees and admires \#Person2\#'s \\
              & dress and outfit. \\ \midrule
              $length=25$ &  \#Person1\# wishes Brian a happy birthday and invites him to dance. Brian agrees and admires \#Person2\#'s \\
              & dress and necklace. They will have a drink together. \\ \midrule
              $length=30$ &  \#Person1\# wishes Brian a happy birthday and invites him to dance. Brian agrees and compliments \#Person2\#'s \\
              &  dress and necklace. They think the party is fine and decide to have a drink together. \\ \midrule
              $length=35+$ & \#Person1\# wishes Brian a happy birthday and invites him to dance. Brian agrees and compliments \#Person2\#'s \\
              &  dress and necklace. They think the party is fine and decide to have a drink together to celebrate his birthday. \\ \bottomrule
            \end{tabular}
            \end{adjustbox}

        \caption{An example of model outputs for length controllability.}
        \label{tab:ctrlen_example}
        \end{table*}

\end{document}